\documentclass[a4paper,conference]{IEEEtran}
\ifCLASSINFOpdf
\else
\fi
\usepackage{epsfig}
\usepackage{subfigure}
\usepackage{calc}
\usepackage{amssymb}
\usepackage{amstext}
\usepackage{amsmath}
\usepackage{multicol}
\usepackage{pslatex}
\usepackage{graphicx}
\usepackage{comment}
\usepackage{color}
\usepackage{multirow}
\usepackage[misc]{ifsym}

\usepackage{todonotes}

\usepackage{todonotes}
\usepackage{bm}
\usepackage{hyperref}

\graphicspath{{img/}}

\DeclareMathAlphabet{\mathcal}{OMS}{cmsy}{m}{n}
\SetMathAlphabet{\mathcal}{bold}{OMS}{cmsy}{b}{n}

\newcommand{\fatz}{\mathbf{z}}
\newcommand{\deltaz}{\Delta \mathbf{z}}
\newcommand{\fatw}{\mathbf{w}}
\newcommand{\deltaw}{\Delta \mathbf{w}}
\newcommand{\fatr}{\mathbf{r}}

\newcommand{\fatq}{\mathbf{q}}
\newcommand{\fats}{\mathbf{s}}
\newcommand{\fatE}{\mathbf{E}}
\newcommand{\fatR}{\mathcal{R}}
\newcommand{\dsdq}{\frac{\partial \mathbf{s}}{\partial \mathbf{q}}}

\begin{document}
%
\title{Generic Merging of Structure from Motion Maps with a Low Memory Footprint}

\author{\IEEEauthorblockN{Gabrielle Flood,
David Gillsj{\"o},
Patrik Persson,
Anders Heyden and
Kalle {\AA}str{\"o}m}
\IEEEauthorblockA{Centre for Mathematical Sciences\\
	Lund University\\
	Lund, Sweden \\
	Email: \{gabrielle.flood, david.gillsjo, patrik.persson, \\anders.heyden, kalle.astrom\}@math.lth.se}}


%


\maketitle
\begin{abstract}
With the development of cheap image sensors, the amount of available image data have increased enormously, and the possibility of using crowdsourced collection methods has emerged. This calls for development of ways to handle all these data. In this paper, we present new tools that will enable efficient, flexible and robust map merging. Assuming that separate optimisations have been performed for the individual maps, we show how only relevant data can be stored in a low memory footprint representation. We use these representations to perform map merging so that the algorithm is invariant to the merging order and independent of the choice of coordinate system.
The result is a robust algorithm that can be applied to several maps simultaneously.
The result of a merge can also be represented with the same type of low-memory footprint format, which enables further merging and updating of the map in a hierarchical way.
Furthermore, the method can perform loop closing and also detect changes in the scene between the capture of the different image sequences.
Using both simulated and real data --- from both a hand held mobile phone and from a drone --- we verify the performance of the proposed method.
\let\thefootnote\relax\footnotetext{\copyright 2021 IEEE.  Personal use of this material is permitted.  Permission from IEEE must be obtained for all other uses, in any current or future media, including reprinting/republishing this material for advertising or promotional purposes, creating new collective works, for resale or redistribution to servers or lists, or reuse of any copyrighted component of this work in other works.}
\end{abstract}


%
\IEEEpeerreviewmaketitle


\section{Introduction}
Over the last couple of years the availability of cheap image sensors --- such as cameras in mobile phones --- has increased immensely. This allows for fast and relatively straightforward collection of large datasets through crowdsourcing. The images can be used to create 3D maps of the environment. However, the more data there are, the heavier the computations for creating these maps will be and due to this, there is a need for faster algorithms for creating 3D maps. Furthermore, additional research on how to fuse individual maps into one global, more accurate map is needed. One use case of such algorithms can be found in the industry for self-driving cars. With a fast and accurate way to merge individual sub-maps, each car that drives in an environment could create its own local map and use that to contribute to a global map. 


\begin{figure}
	\centering
	\includegraphics[width=0.32\textwidth, trim = 0cm 0cm 0cm 0cm]{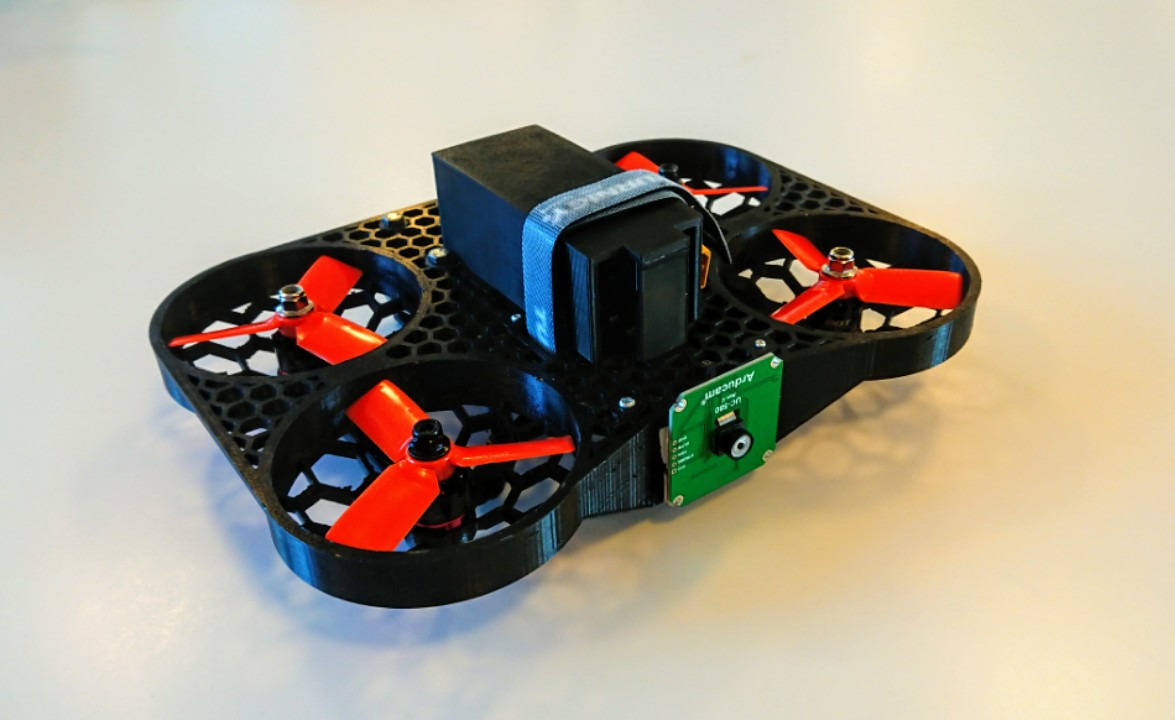}
	\vspace{-3mm}
	\caption{A drone equipped with a camera, IMU and a Raspberry Pi. The drone is used for real-time mapping and was used for one of the real data experiments in this paper.}
	\label{fig:drone}
\end{figure}

Estimating map parameters and sensor motion using only sensor data is referred to as simultaneous location and mapping (SLAM) \cite{slamref,davison-et-al} and structure from motion (SfM) \cite{hartley2003multiple}. Classically, SLAM has focused more on the motion, while SfM has been more focused on the structure. Also, SLAM often requires that one moving camera is used, while SfM can be used for unsorted images from different cameras. Nevertheless, the two methods essentially solve the same problem, but are originating from different research fields. 

When image data are used, the SfM is usually performed using \textit{bundle adjustment}. The name refers to the bundle of rays going from each 3D point in space to each camera and it can be seen as a large sparse geometric parameter estimation problem, cf.\ \cite{triggs1999bundle}. Bundle adjustment is commonly used both as a final step and as an intermediate step in the optimisation to prevent error buildup \cite{engels2006bundle,klein2007parallel}. It can also be used to merge maps, by doing a new optimisation over all data at once. However, bundle adjustment is a computationally expensive process and there is a need for making these methods more efficient.

A faster method to align two maps is to use point cloud registration. One example of a commonly used method for registration is iterative closest point (ICP) \cite{chen1992object}. This does not require any knowledge of point matches between the different sets. If such matches are known, one can instead use e.g.\ Procrustes analysis \cite{kendall1989survey}. The methods for point cloud registration do not, however, solve the merging problem, but leaves a map with double representations of matching points.

When it comes to fusion of individual maps, there are different ways to do this, but many of the methods are developed for concurrent mapping. Several of them have been created to perform collaborative visual SLAM. There are examples of collaborative visual SLAM that work for several units at once and are fast enough to run in real-time \cite{davison2007monoslam}. Many of these examples are focused on implementations in drones flying simultaneously. In these cases, parts of the pipeline are run on the platform, while parts are computed in the cloud. The map fusion is then based on a few keyframes, to decrease the need for storage space \cite{schmuck2017multi}. There are also several examples where the bundle adjustment is only performed locally to decrease the computational effort \cite{mouragnon2006real}.
In the collaborative SLAM method presented by \cite{zou2013coslam} a dynamic environment is possible and several cameras can be used. However, the cameras are initialised by viewing the same scene. This simplifies the global coordinate system, but is not always applicable, since it is often the case that there are no common points for all bundle sessions. There are also studies where several cameras have been used at the same time, but when they are fixed on a stereo head \cite{nister2004visual}. All these methods are developed for simultaneous mapping using several cameras. When maps from different occasions are merged the conditions change, which gives other limitations and possibilities.

Another important problem within SfM is the ability to perform \textit{loop closure}.
This problem appears when a reconstruction is made iteratively on a long image sequence and some feature points reappear after some time.
Due to the inherent drift and error accumulation the reappearing points will not be reconstructed at the same position as they where reconstructed initially.
For the loop closure problem it is assumed that it is possible to identify which points in the images that belong to the same 3D point.
When re-appearing points are detected, it is possible to utilise this information and increase the quality of the reconstruction and at the same time position these at the same 3D location.
Some techniques for loop closure can be found in 
\cite{newman,Williams-et-al,guilbert-kahl-etal-accvj-04}.

There are also examples where SLAM is solved using a Bayesian approach \cite{davison2003real}, which is faster but not as accurate as bundle adjustment \cite{triggs1999bundle}.
The methods that are discussed so far in this paper are not only applicable to images, but work similarly for other sensor data as well, e.g.\ wifi \cite{batstone2016robust} and audio \cite{kuang2013complete,zhayida2014automatic}. Gaining information and ideas from these fields can thus be useful for SfM as well.

In \cite{flood2019merging}, a method that is a compromise between a full optimisation bundle and the Kalman filter was presented. The method was primarily evaluated on audio data together with a small experiment for image data. In this paper we develop that idea further to work automatically for SfM data from RGB images. The idea behind the method presented in \cite{flood2019merging} is that maps can be merged efficiently using only a small memory footprint from the map and the residuals. Then the merging problem can be solved linearly. For this to work on images from different datasets there is a need for a coordinate system estimator. 

In this paper, we present a method for efficient and simultaneous estimation of the parameters --- i.e.\ camera matrices and 3D points --- as well as the coordinate system. The system is also adopted for partially non-overlapping data. 
The pre-process of the data starts with the detection of feature point descriptors, e.g.\ using SIFT \cite{lowe2004distinctive} or ORB \cite{rublee2011orb}, whereupon the individual maps are estimated using a SfM pipeline \cite{olsson2011stable}. The data can be collected using a hand-held camera or an autonomous drone, like the one in Figure~\ref{fig:drone}. The individual maps are then fused at the same time as transformations into a global coordinate system are estimated using only the map points and a compressed representation of the Jacobian. This means that once the individual maps are computed, there is no need for saving the actual images --- not even a few keyframes. 

The main contribution of this paper is the generic and efficient method for merging sub-maps obtained from several image sequences. The proposed method is independent of the chosen order of the sequences and the choice of coordinate system. It is furthermore very efficient compared to making a full bundle adjustment of all image sequences together, by utilising a compact and efficient representation of each bundle, consisting of a considerably reduced number of free variables. The merging method can also be used to detect changes in a scene and to solve the loop closure problem. This is validated on both simulated and real data.

\section{SfM Systems and Bundle Adjustment}
The pre-processing steps, such as creating the individual map representations, are not the focus of this paper. Nonetheless, we will briefly go through the theory. Many of the notations that will be used later in the paper are introduced in this section. The purpose of the individual optimisation bundles is to find the $m$ camera matrices $P_i$ and the $n$ 3D points $U_j$ that induce the image points $u_{ij}$. Each image point gives rise to two residual terms $r_{ij}$, one for each image coordinate, when it is compared to the projection of $U_j$ in camera $i$,
\begin{equation}
\label{eq:residual}
r_{ij} = \begin{bmatrix}
\frac{P_i^1 U_j}{P_i^3 U_j} - u_{ij}^1 \\
\frac{P_i^2 U_j}{P_i^3 U_j} - u_{ij}^2
\end{bmatrix} .
\end{equation}
Above, $P_i^k$ denotes row $k$ of camera matrix $P_i$ and  $u_{ij}^k$ denotes element $k$ of $u_{ij}$. The total residual vector $\fatr$ is composed by stacking all individual residual vectors $r_{ij}$. Furthermore, we collect the unknown parameters in a structure $\fatz$, s.t.\
\begin{equation}
\fatz = (P_1,\, P_2, \, \hdots \, P_m, \, U_1, \, U_2,\, \hdots \, U_n) .
\end{equation}

In the optimisation we use local parametrisations, $\deltaz \in \fatR^{6m+3n}$, around each point $\fatz_0$ in the parameter space,
\begin{equation}
(\fatz_0,\deltaz) \longrightarrow \fatz .
\end{equation}
While $\fatz$ contains twelve parameters for each camera, the local parametrisations only use six parameters per camera, in order to assure that the camera is composed by a rotation matrix and a translation matrix. To find out how a change $\deltaz$ affects the residual $\fatr$, we compute the derivatives of the components in $\fatr$ with respect to the elements in $\deltaz$. Even if we only refer to the six parameters per camera together with the 3D points we will for simplicity further on denote this Jacobian $J=\partial \fatr / \partial \fatz$. The maximum likelihood estimate $\fatz ^{\ast}$ of $\fatz$ is found by minimising the sum of squared residuals,
\begin{equation}
\label{eq:delta_z}
\mathbf{z}^* = \text{argmin}_{\mathbf{z}} \fatr^T \fatr.
\end{equation}
Using Gauss-Newton, each step of the iterative bundle adjustment corresponds to the parameter update
\begin{equation}
\label{eq:z_update}
\deltaz = -(J^TJ)^{-1}J^T \fatr.
\end{equation}

Performing the optimisation on $N$ separate data collections results in $N$ different parameter representations $\fatz ^{(k)}$, where superscript index $(k)$ denotes the representation number. Some of the 3D points are visible in several representations while some are visible in only one. Note that the ordering might differ, such that $U_{j}^{(k)}$ does not represent the same point as $U_j^{(l)}$. Once  matches between the different data collections have been established, e.g.\ using ORB or SIFT features, the individual map representations can be merged into one global map. One way to do this would be to perform a bundle adjustment with all data from all data collections, but this could be prohibitively expensive in terms of memory and computations. Another way could be to do co-registration of the point clouds, e.g.\ using Procrustes. The naive way to merge the maps would be to then take the average position of matching points. One drawback of this merging method is that the resulting global map is depending on the merging order.

\subsection{A Compact and Efficient Model for a Bundle Session}
\label{sec:medium}
The proposed method exploits the fact that the optimal residuals from the separate bundles can be linearised to avoid the large bundles. Our bundle representation is built on theory from \cite{flood2019merging} and for completeness, we will summarise some of that theory in this section. 

A key idea is to divide the unknown parameters in $\fatz$ into two parts $\fatq$ and $\fats$, where $\fatq$ contains the parameters that potentially could match to those of other SfM sessions. The parameters in $\fats$ can be thought of as auxiliary parameters. There is an interesting trade-off here. Making $\fatq$ larger allows for a higher number of potential matches with other SfM sessions, but requires a large memory footprint and vice versa. In this paper we use the approach that some (or all) of the 3D points go into $\fatq$, whereas the rest of the points and camera matrices go into $\fats$. 

\subsubsection{Approximating the Residual}
The parameters in $\fatz$ are ordered such that $\Delta \fatz = \begin{bmatrix} \Delta \fatq & \Delta \fats \end{bmatrix}^T $. The Jacobian $J$ is divided correspondingly, with the part that corresponds to the parameters in $\fatq$ denoted $J_a$ and one that corresponds to the parameters in $\fats$ denoted $J_b$. The auxiliary parameters in $\fats$ will depend on the points in $\fatq$ as follows
\begin{equation}
\label{eq:dsdq}
\frac{\partial \fats}{\partial \fatq} = -(J_b^TJ_b)^{-1}( J_a^TJ_b)^T.
\end{equation}

That derivative can furthermore be used to express how the residuals change if the points in $\fatq$ are moved. We have that
\begin{equation}
\Delta \fatr =
\Big ( \underbrace{ J_a+ J_b\cdot \dsdq }_{J_q} \Big ) \Delta \fatq .
\end{equation}

Furthermore, viewing the residual as a function of an update $\Delta \fatq$ and linearising it around an optimal point $\omicron$ gives the following approximation of the squared residual
\begin{equation}
\label{eq:squared_residual_3}
\fatr^{T}\fatr \approx a^2 +\Delta \fatq^T R^TR \Delta \fatq,
\end{equation}
where $a^2=\fatr \rvert_{\omicron}^T\fatr_{\omicron}$ and $R$ is a triangular matrix originating from QR-decomposition of  $J_q\rvert_{\omicron}$.

Another way to view this is to form a modified residual vector $\hat{\fatr}$ according to
\begin{equation}
\label{eq:squared_residual_3b}
\hat{\fatr} =
\begin{bmatrix}
a\\
R  \Delta \fatq
\end{bmatrix} =
\begin{bmatrix}
a\\
R  (\fatq - \fatq\rvert_{\omicron})
\end{bmatrix},
\end{equation}
whose sum of squares is an approximation of the original sum of squares, i.e.\
\begin{equation}
\label{eq:approx_error}
\fatr^{T}\fatr \approx \hat{\fatr}^T \hat{\fatr} .
\end{equation}
The linearisation decreases the memory footprint substantially compared to the original problem.

To summarise the theory from \cite{flood2019merging}, the compressed representation of data consists of $(\fatq\rvert_{\omicron}, a, R)$, where $\fatq\rvert_{\omicron}$ is a subset of the 3D points. Note that while $J_q$ is a rectangular matrix, $R$ will be quadratic and thus much smaller than $J_q$. Despite this, it was shown in \cite{flood2019merging} that it is possible to obtain a good approximation of the residual according to Equation~\eqref{eq:squared_residual_3}. Furthermore, once an update has been made, the rest of the points and the camera matrices can be updated using $\partial \fats / \partial \fatq$.

\subsection{Gauge Freedom}
SfM estimates can only be determined up to an unknown choice of coordinate system, which is called gauge freedom. This involves translation, rotation and change of scale, which in total has seven degrees of freedom. A consequence of this is that the Jacobian $J_q$ and also the matrix $R$ has a seven-dimensional nullspace. The process of changing coordinate system is however non-linear, and therefore the approximation we presented in the previous section is only valid for points close to the optimal point $ \fatq\rvert_{\omicron} $. In \cite{flood2019merging} the different maps were pre-aligned and the gauge freedom was therefore less relevant.

\section{Merging Several SfM Sessions}
As we mentioned before, one way to add several individual SfM sessions would be to do a new bundle, over all data. However, this could be computationally expensive and require storing a large amount of data. The faster approach presented in \cite{flood2019merging} solved the problem linearly. Though, this did require that the individual map representations were aligned and that a number of points were visible in all maps. In this section, we generalise this further to work for any representations and handle the coordinate ambiguity. Thus, no pre-alignment  is needed and that makes the approach much more flexible.

Assume that for each map $k$ we have the compressed information as $(\fatq ^{(k)},a^{(k)},R^{(k)})$. Denote the global map $\fatq$ and let that contain some or all of the 3D points that are contained in at least one of the individual maps $\fatq ^{(k)}$. 

If $\fatq$ is assumed to contain $\bar{n}$ 3D points, all potential global map representations lie on a $3\bar{n}$-dimensional manifold. Each representation can then be projected to lower dimensional spaces in which the local map representations lie. In practice, the projection $p_k(\fatq)$ simply means that we leave some points out, while we keep the rest, i.e.\ $p_k(\fatq)$ has the same number of points (and the same point order) as $\fatq^{(k)}$. However, they might be in a different coordinate system. Therefore, we apply a similarity transform $T_k$ to obtain a local map representation. Hence, we would like $T_kp_k (\fatq)$ to be close to $\fatq^{(k)}$.

Thence, the unknowns are the global map  $\fatq$ and the $N$ different transformations $T_k$. By collecting individual residuals, similar to the ones in \eqref{eq:squared_residual_3b}, the approximate modified residuals
\begin{equation}
\label{eq:error_function_merge}
\hat{\fatr} =
\begin{bmatrix}
a^{(1)}\\
R^{(1)}  (T_1p_1 (\fatq) - \fatq^{(1)}) \\
\vdots \\
a^{(N)} \\
R^{(N)}  (T_Np_N (\fatq) - \fatq^{(N)})
\end{bmatrix}
\end{equation}
are such that $\hat{\fatr}^T \hat{\fatr}$
is approximately equal to the sum of the squared residuals for all $N$ sessions.

%

Our approach to solve this problem is to bundle over $\fatq$ and all $T_{k}$, trying to minimise \eqref{eq:approx_error}
with  $\hat{\fatr}$ according to \eqref{eq:error_function_merge}.
This bundle will be significantly smaller and faster than bundling over all the original images with the re-projection errors as loss.

\begin{figure}
	\centering
	\includegraphics[width=0.35\textwidth, trim = 4cm 4cm 10.5cm 2cm]{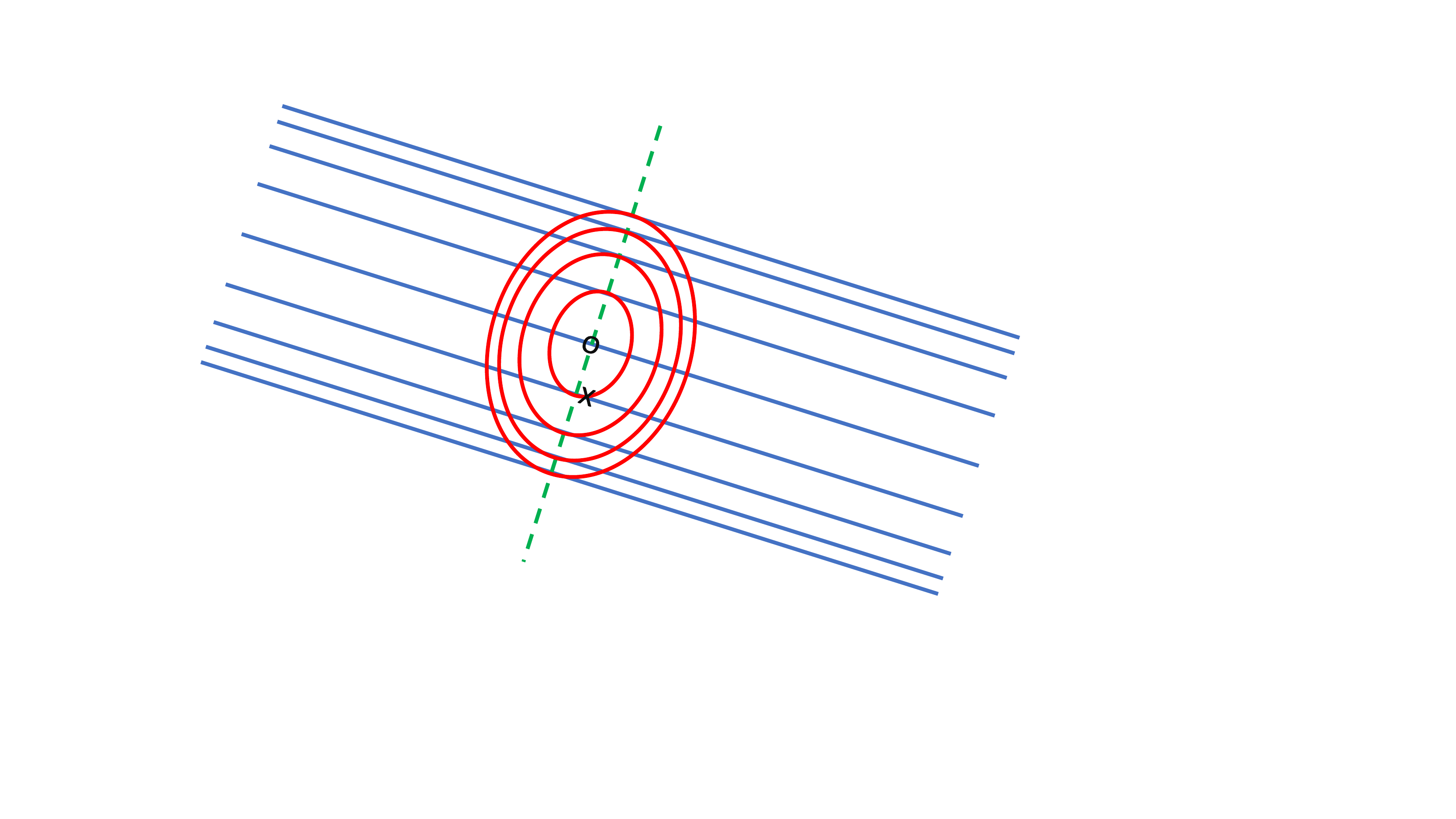}
	\vspace{-5mm}
	\caption{The blue solid lines show the level curves of the linearised error function and the o shows the point we have linearised around. By adding a penalty in the perpendicular direction, along the dashed green line, the resulting error function is the one shown as red ellipses.}
	\label{fig:error_function}
\end{figure}

One difficulty with this approach is that the approximation \eqref{eq:squared_residual_3} only holds when each map $T_kp_k (\fatq)$ is close to its working point $\fatq^{(k)}$. An interesting thing to note here is that the last seven rows of $R^{(k)}$ --- the triangular matrix which comes from QR-decomposition of the Jacobian w.r.t.\ $\fatq$ of bundle $k$ --- will be zero, due to the gauge freedom.
An illustration of this is given in Figure~\ref{fig:error_function}, where we visualise this in a lower dimension. The error function locally looks like a parabolic cylinder (illustrated with blue level curves in the figure). To force the error function to be quadratic --- a paraboloid --- we change the last seven rows of $R^{(k)}$ such that they are orthogonal to the rest of the rows (see the green dashed line). The change of the last rows of $R^{(k)}$ results in an error function that has the level curves shown by the red ellipses. By optimising over the transformation $T^{(k)}$, we will end up at a point where the orbit is tangent to a level curve (a red curve in Figure~\ref{fig:error_function}). One such point could be the one marked by x in the image. Since this x is close to our working point o, the linearisation is still valid and the addition of the last rows of $R^{(k)}$ will have little undesired effect.

\subsection{The Bundle Initialisation}
The initialisation for the merge bundle can be done in several ways. 
We have decided to initialise $\fatq $ from $ \fatq^{(1)}$, and $T_{1}$ to be the identity matrix, while we initialise the rest of the $T_{k}$:s using Procrustes analysis between matching points in $\fatq^{(k)}$ and $\fatq ^{(1)}$.  Points in $\fatq$ that are not in $\fatq^{(1)}$ are initialised from the other maps and $T_{k}$.
Initial estimates between two maps can be obtained using three point correspondences, but there are also solvers for mini-loop closure involving fewer than three points between three, four and five maps, cf.\ \cite{miraldo2019minimal}.

\subsection{The Bundle for Merging the Maps}

Similar to the individual bundle approach, we collect the unknown variables in a structure $\fatw$, s.t.\
\begin{equation}
\fatw = ( \fatq, \, T_1,\, T_2, \, \hdots \, ,T_N),
\end{equation}
and use local parametrisations, $\deltaw \in \fatR^M$, around each point $\fatw_0$ in the parameter space,
\begin{equation}
(\fatw_0,\deltaw) \longrightarrow \fatw .
\end{equation}
The dimension $M$ of $\deltaw$ depends on how many of the points in $\fatq$ that are common between the individual maps. In the local optimisation we use a Levenberg-Marquart approach. In each step we calculate the Jacobian $J_w$ that describes how changes in the parameters $\deltaw$ affect the residual $\hat{\fatr}$.

\subsection{Compressing the Result from the Merge}
Once the merge is done, the residuals can be compressed for future merges.
The Jacobian $J_w$ is again divided in one part $\bar{J}_a$ that corresponds to the parameters in $\fatq$ and one $\bar{J}_b$ that corresponds to the rest of the parameters $\fats$ (now corresponding to changes in $T_1, \ldots, T_N$ and some of the 3D points).
From this we again calculate how $\fats$ depends on $\fatq$, similar to what we did in Equation~\eqref{eq:dsdq}.
We then calculate $\bar{R}$ from a QR-factorisation of  $\bar{J}_q = \bar{J}_a+ \bar{J}_b\cdot \partial \fats/\partial \fatq$.
In this way, a compact representation $(\fatq,\bar{a},\bar{R})$ of the result can be calculated, again similar to what we did for the individual bundle sessions. The value $\bar{a}^2$ is the squared residual in the optimal point.

\section{Hypothesis Testing of the Merge} \label{sec_hyp_testing}
Even if we assume that the matches between the maps are given, some of them might be wrong. Also, there could be other errors in the merge, e.g.\ if any object in the scene has moved. For this reason, some hypothesis test is needed. We use the same approach as in \cite{flood2019merging} and compare the increased error to a $\Gamma$ distribution, and extend that theory here. Again, let $N$ denote the number of map representations that are merged.

If we assume that the measurement errors in the images are zero mean Gaussian with a standard deviation $\sigma$, the expected value of the squared residuals  $(a^{(k)})^2$ in the individually optimal points from \eqref{eq:squared_residual_3} are
\begin{equation}
\label{eq:ak2}
\fatE [ \big( a^{(k)}\big)^ 2 ]= \fatE [ \big( \fatr^{(k)} \big)^T \big( \fatr^{(k)} \big) ] =
\sigma^2\big(\eta _{res}^{(k)} - d_{dof}^{(k)}\big).
\end{equation}
We denote the number of residuals in bundle $k$ by $\eta_{res}^{(k)}$ and the effective degrees of freedom $d_{dof} ^{(k)}$. In a bundle with $m$ cameras and $n$ 3D points, these will be $\eta_{res}=2mn$ and $d_{dof}=6m+3n-7$, where the $7$ represents the gauge freedom. Furthermore, if we assume that the merge was successful, the expected value for the merged map will be
\begin{equation}
\fatE [ \bar{a}^ 2 ]= \fatE [ \fatr^T \fatr ] =
\sigma^2(\eta _{res}-d_{dof}),
\end{equation}
where $d_{dof}$ is the effective degrees of freedom in the merge and $\eta _{res}$ is the total number of residuals. We have that $\eta _{res}=\sum_k\eta _{res}^{(k)}$, while the value of $d_{dof}$ will depend on the overlap between the individual map representations.

Now, for the difference between $\bar{a}^ 2$ and all $(a^{(k)})^2$, we have
\begin{equation}
\fatE \Big[ \underbrace{ \bar{a}^ 2 - \sum_k \big( a^{(k)}\big)^ 2 }_{\tilde{a}}\Big] =
\sigma^2 \Big(\eta _{res}-d_{dof}- \sum_k \Big( \eta _{res}^{(k)} - d_{dof}^{(k)} \Big) \Big).
\end{equation}
Letting  $\tilde{a}= \bar{a}^ 2 - \sum_k \big( a^{(k)}\big)^ 2$  and denoting the number of 3D points that are common in $i$ individual maps $\kappa_i$, this gives
\begin{equation}
\label{eq:a2_diff}
\begin{aligned}
\fatE [ \tilde{a}] =&
\sigma^2 \Big( \sum_k  d_{dof}^{(k)} -d_{dof} \Big) \\
=&\sigma^2 \left( \Big( \sum_{i=1}^N 3\kappa_i(i-1) \Big)  -7\cdot(N-1) \right),
\end{aligned}
\end{equation}
where the factor $3\kappa (i-1)$ represents that we have locked another $3\kappa_i$ point coordinates $i-1$ times. We subtract by $7\cdot(N-1)$ since all individual maps are now merged to the same coordinate system.

Furthermore, since the noise is Gaussian, the value in \eqref{eq:a2_diff} will be a sum of $\sum_k  d_{dof}^{(k)} -d_{dof}$ Gaussian distributed variables. Altogether, this means that for a successful merge, $\tilde{a}$ should come from a $\Gamma$ distribution with the following density \cite{feller1966introduction}
\begin{equation}
\label{eq:gamma_distribution}
f_{\alpha,\nu}(x) =\frac{1}{\Gamma(v)}\alpha^\nu x^{\nu-1}\text{e}^{-\alpha x}.
\end{equation}
Here,  $\Gamma$ is the gamma function and
\begin{equation}
\alpha=\frac{1}{2\sigma ^2}, \quad  \nu=\frac{\big( \sum_{i=1}^N 3\kappa_i(i-1)\big)  -7\cdot(N-1)  }{2}.
\end{equation}
Hence, if the merge for some reason was not successful, this could be discovered by comparing the value of $\tilde{a}$ to the expected $\Gamma$ distribution. Using this, we can detect whether changes has occurred in the scene between the different mapping occasions.

The standard deviation $\sigma$ of the noise is often unknown. Nevertheless, it can be estimated as the mean of the standard deviations for the individual map representations, which in turn would be estimated according to \cite[p. 47]{hastie2009elements}.

\section{Using Merging for Increased Robustness}
Once we know which $\Gamma$ distribution the increased error $\tilde{a}$ should come from, this can be used as a hypothesis test. This could furthermore be used to increase robustness in a large SfM session. If we do a bundle over a scene and the residuals are not sufficiently small, one might suspect that there is corrupt data or outliers involved. The SfM session could then be divided to a number of parts, where SfM first is performed on each of them, resulting in a number of sub-maps. Given that the different sub-maps are divided such that they have overlap, they can be merged using our method. By merging one part at a time and checking the distribution of $\tilde{a}$, it can be found where in the dataset there is corrupt data. That part can thereafter be divided into smaller parts, and the process can be repeated. We can by that avoid to add erroneous information to the global map, while we successfully can add the other parts which are correct.
%

\section{Experimental Validation}
To verify the proposed method we have run a number of experiments, both on simulated and real image data. The experiments are described in an order of increased complexity.

\subsection{Verification on Simulated Data}
First, we verified the method and the hypothesis test on simulated data. We simulated $100$ 3D points $U_j$ in a box of size $10\times 6 \times 2$ and ten cameras $P_i$ pointing towards the box. The cameras were re-simulated three times to mimic three mappings. All 3D points were visible in all cameras and we added Gaussian noise with zero mean and standard deviation $\sigma = 0.05$ in the image projections. On each mapping we separately performed bundle adjustment using only the image projections $u_{ij}$, resulting in three different representations of the same scene, each given in a different coordinate system. We found matches between all three map representations and using ten of these matches in $\fatq$, we merged the maps into one global map.

\begin{figure}
	\centering
	\includegraphics[width=0.32\textwidth, trim = 0cm 0cm 0cm 0cm]{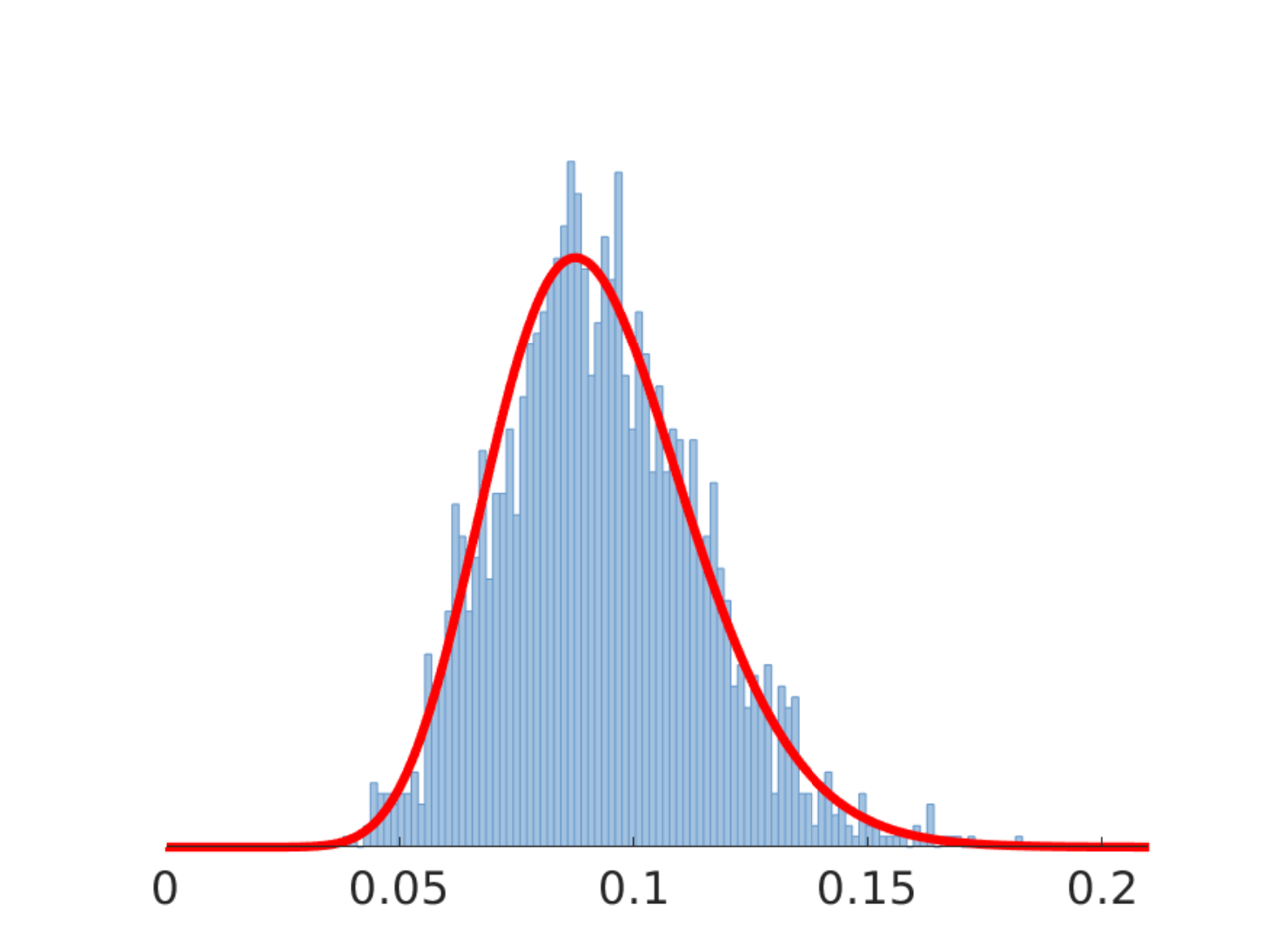}
	\vspace{-3mm}
	\caption{The image shows the histogram of $\tilde{a}$ achieved from running the same experiment several times with different noise realisations. The histogram is expected to follow the $\Gamma$ distribution shown by the red curve.}
	\label{fig:histogram}
\end{figure}

We repeated the experiment above $2\,000$ times with the same cameras and 3D points, but with different noise realisations. For each run, we saved $\tilde{a}$ and finally we plotted a histogram over the result. Figure~\ref{fig:histogram} shows the histogram together with the expected $\Gamma$ distribution from Equation \eqref{eq:gamma_distribution}. The figure clearly shows that the error follows the distribution even though we have linearised the residual according to Equation \eqref{eq:error_function_merge}. This is a verification that the method works even when we are optimising over the transformations to the global coordinate system as well as the global map.

\subsubsection{Early Stopping of Pre-Processing}
\begin{figure}
	\centering
	\includegraphics[width=0.32\textwidth, trim = 0cm 0cm 0cm 0cm,clip]{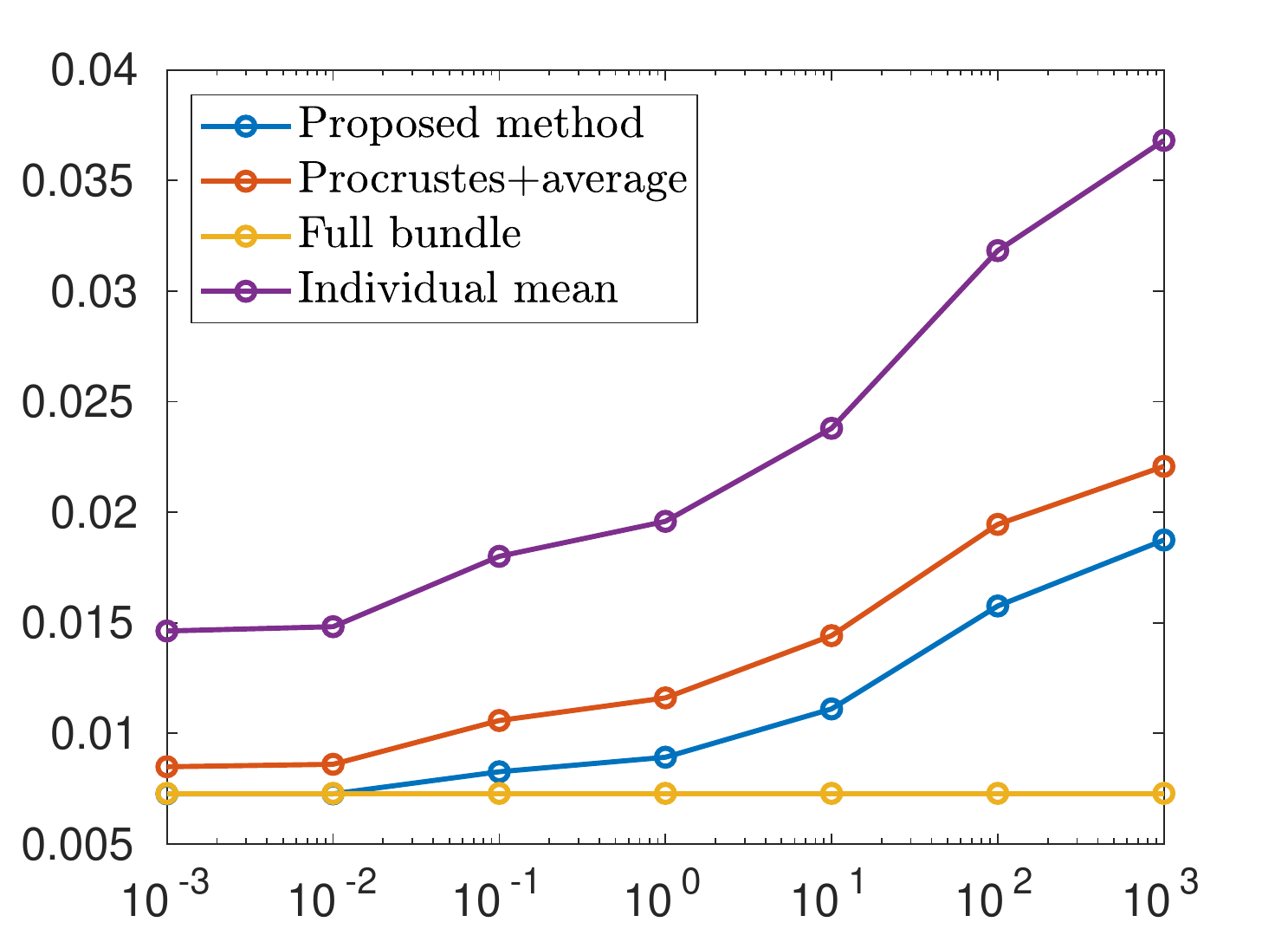}
	\vspace{-3mm}
	\caption{How the RMSE for the final map achieved using different merging methods change when the individual map bundles are terminated at different levels. The x-axis shows the Euclidean norm of the gradient and the y-axis the RMSE. The error for the individual maps is included for comparison.}
	\label{fig:earlystop}
\end{figure}

In the previous experiment we let the individual map representations reach an optimal state before merging them. However, this is not always the case in reality, due to poorly chosen bundle thresholds or to shortage of time. In the second experiment we investigated how the system performance degrades with less optimisation in the pre-processing steps. The setup was similar; all 3D points were visible in all map representations (but only in 80 \% of the cameras). We used noise with $\sigma = 0.005$. We stopped the bundle for the individual maps based on the Euclidean norm of the gradient $2\hat{\fatr}J$, which is obtained by differentiating \eqref{eq:approx_error}, with the residual given by \eqref{eq:squared_residual_3b} and the Jacobian $J$ defined from that. The termination was set at different levels, after which we used our proposed method for merging. For comparison, we performed a large bundle on all data, and we also did Procrustes registration (using an arbitrary order) followed by averaging over matching points. The mean RMSE over 1\,000 runs was computed and is plotted in Figure~\ref{fig:earlystop}. For comparison, the mean RMSE for the three individual map representations is shown as well. To compute the RMSE we first did Procrustes registration of the respective map to the true map. 

Our proposed method performs better than Procrustes at all stages, and furthermore the graph is less steep than that of the individual errors, which means that some of the performance that is lost from the early stopping is recovered using our method. Finally, one can see that for small gradient norms, our method performs as well as the large bundle, which is much more computationally expensive.

\subsubsection{Solving Loop Closure by Map Splitting}
Furthermore, we wanted to show that our method can solve the problem of loop closure, not within one individual bundle, but in the merging of several slightly overlapping sessions. We also decreased the number of parameters in $\fatq$ to be a small part of all the 3D points. First off, we simulated a SfM session of a room of size $5\times 6 \times 2$ m and divided it into four parts, such that each sub-map captured one of the walls, with a few corner points common between the different sub-maps, and no points common in more than two maps. Each sub-map consisted of $200$ 3D points and only 6 \% of these coincided with points from any of the other maps. The noise level was $\sigma = 0.005$.

\begin{figure}
	\centering
	\hspace{-1cm}
	\includegraphics[width=0.27\textwidth, trim = 0cm 0cm 0cm 0cm]{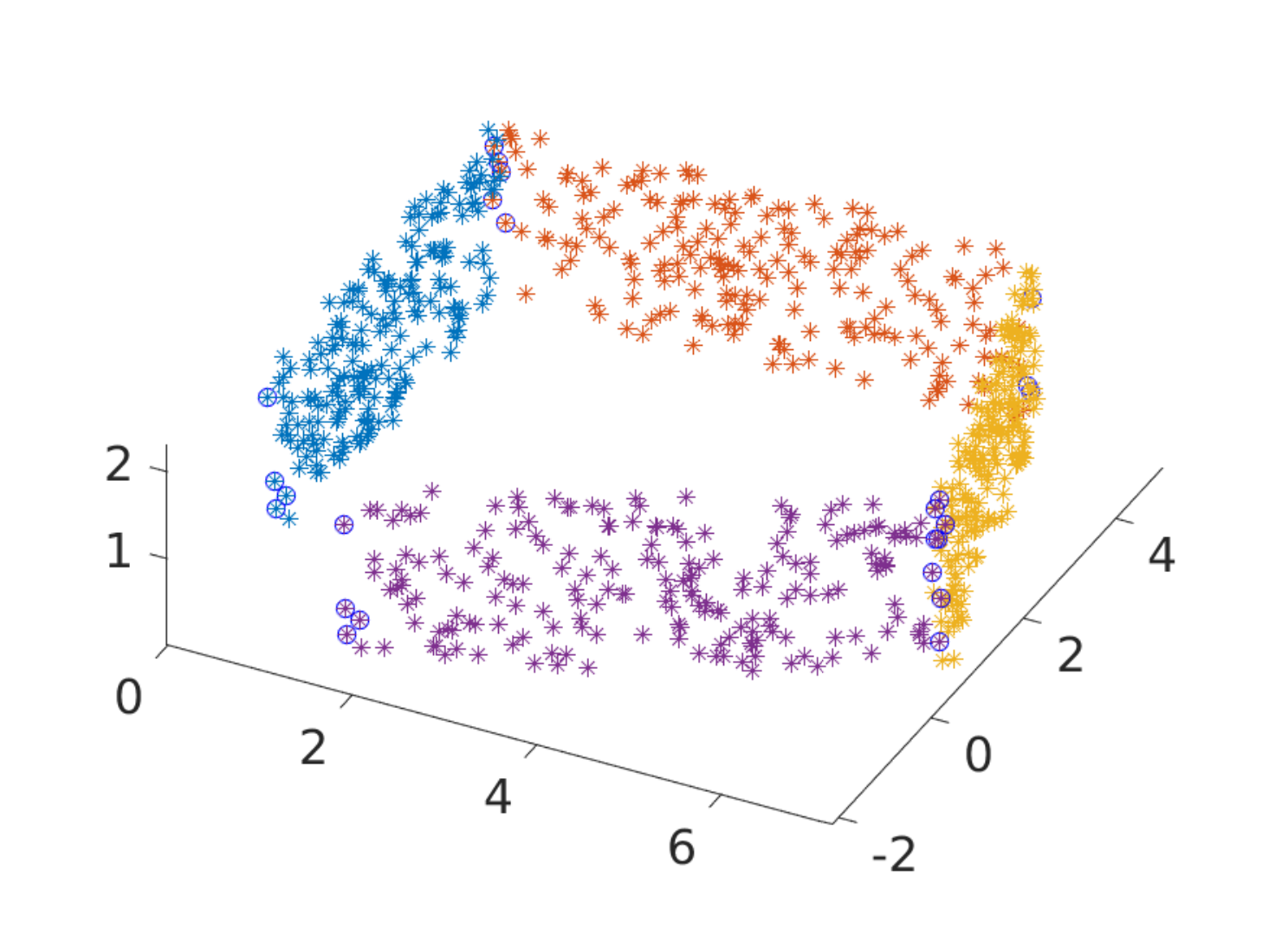}
	\hspace{-1cm}
	\includegraphics[width=0.26\textwidth, trim = 0cm 0cm 0cm 0cm]{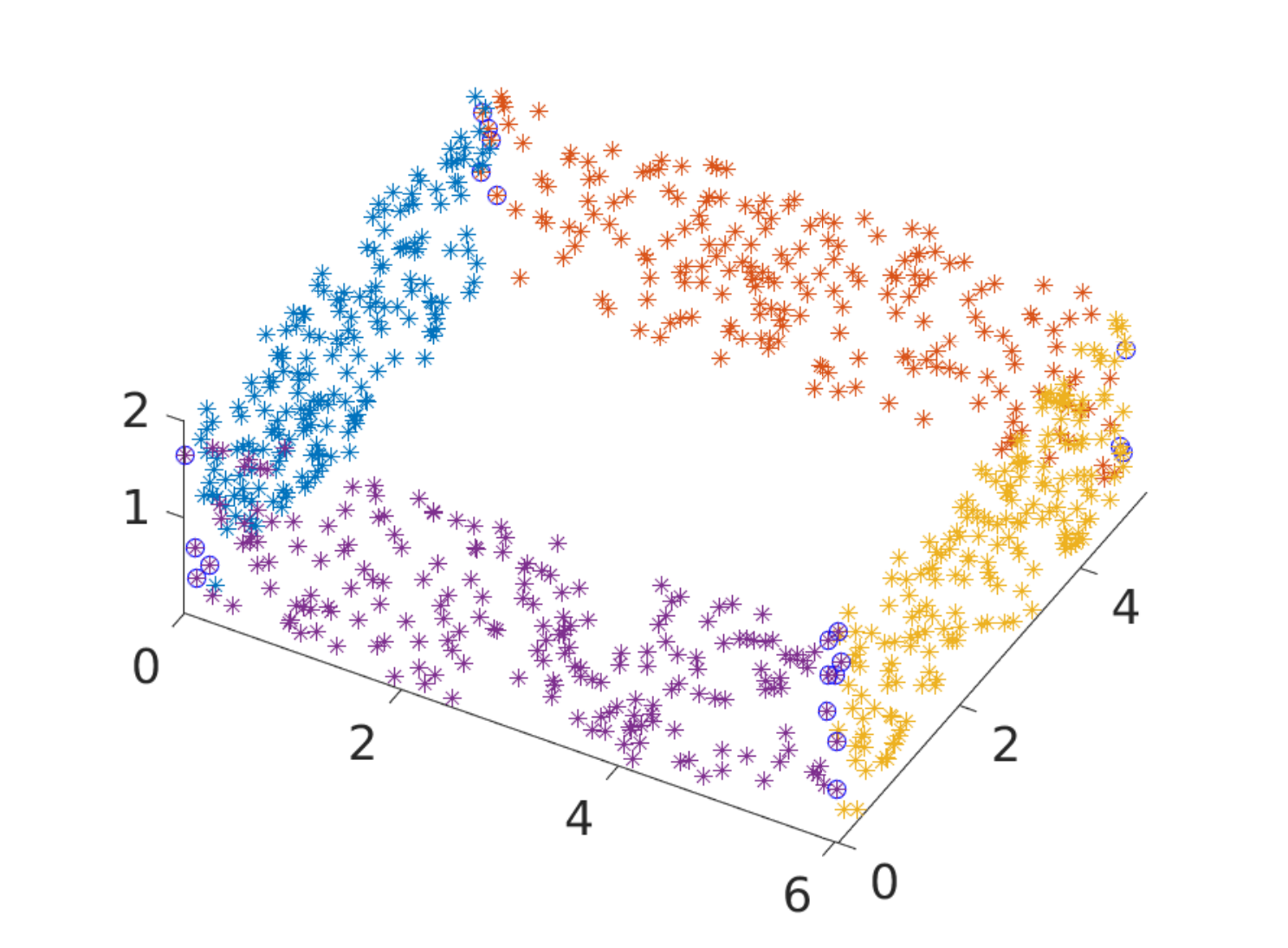}
	\hspace{-1cm}
	\vspace{-3mm}
	\caption{The left plot shows the result from merging the different sub-maps without matches in the beginning and the end of the map. To the right we have added matches between these two and the loop closure problem is solved.}
	\vspace{-1mm}
	\label{fig:loop_closure}
\end{figure}

To simulate a loop closing problem we used two merging methods --- one with matches between sub-maps 1-2, 2-3 and 3-4 and one where there were matches between sub-maps 1-4 as well. The first case represents what happens when you do SfM starting at one point of the room and do a loop without using any loop closing technique, while the latter uses our proposed method. We could see that our method improves the performance concerning loop closure. In most cases there was a drift in the map for the first method, but if we added matches between sub-maps 1-4 as well, this drift disappeared. Figure~\ref{fig:loop_closure} shows how the method fails to connect the ends of the blue and the purple sub-maps in the first case, but succeeds in the second. 

Running the same experiment $1\,000$ times shows that adding matches between sub-maps 1-4 gives a reduced Euclidean distance from the ground truth in 85 \% of the cases, and the distance is reduced by 50 \% or more in 80 \% of the cases. In terms of RMSE, this error was less than $0.1$ in 99.8 \% of the cases for the full bundle. This can be considered as gold standard. The corresponding value for our method with all matches was 80 \%; for our method without 1-4 matches 25 \%; and for Procrustes and averaging 7.7 \%. The mean RMSE within those 80 \% for our method was 0.038. If the merge is unsuccessful, the RMSE value is not very suggestive, since the registration made for comparison might be wrong too.

Furthermore, Table~\ref{tab:loop_closure} illustrates how much the memory footprint is decreased when we use our compressed error representation. The linearised residual in \eqref{eq:squared_residual_3} reduces the parameters in the Jacobians from approximately $3\, 000 \times 660$ to $30 \times 30$ compared to the full residual. This becomes even more evident when we look at the size of the bundle for the map merging, compare Equations \eqref{eq:approx_error} and~\eqref{eq:error_function_merge}. All this show that our proposed method performs best except for the full bundle and that it therefore is a very good compromise between performance and efficiency. 

\begin{table}
	\caption{Four sub-maps were merged. The table shows the size of the Jacobian for using the full residual \eqref{eq:residual} and the linearised residual \eqref{eq:squared_residual_3} for each sub-map. The last line shows how much smaller the merging problem becomes using our method.}
		\label{tab:loop_closure}
	\centering
	\begin{tabular}{lllll}
	Bundle	&\# points & Size of full& Size of  compressed\\
	session	&	&  Jacobian &   Jacobian \\
	\hline
	1	&	 200 & $3\, 082 \times 660$ & $27 \times 27$ \\
	2	&	 200 & $2\, 792 \times 660$ & $24 \times 24$ \\
	3	&	 200 & $3\, 140 \times 660$ & $33 \times 33$ \\
	4	&	200 & $3\, 190 \times 660$ & $36 \times 36$ \\
	merge	&	784 & $12\, 204 \times 2\, 601$ & $120 \times 88$ \\
	\end{tabular}
\end{table}

\subsection{Verification on Real Data}

\subsubsection{Small Bookshelf Experiment}
\begin{figure}
	\centering
	\includegraphics[width=6cm]{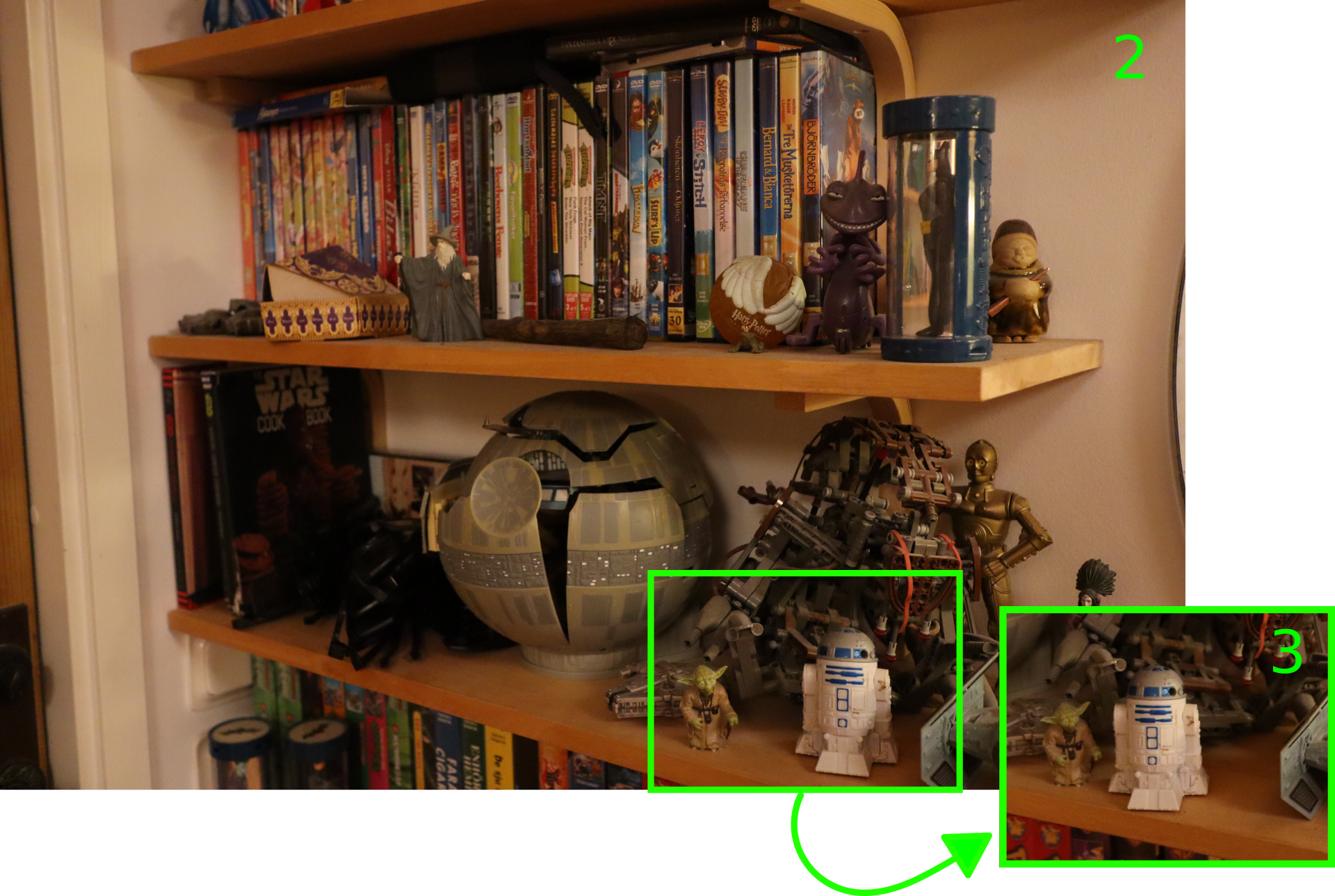}
	\vspace{-3mm}
	\caption{This figure shows how the R2D2 model moved between collection 2 and 3 in the bookshelf experiment.}
	\label{fig_r2d2}
\end{figure}

\begin{figure}
	\centering
	\includegraphics[width=0.32\textwidth, trim = 0cm 0cm 0cm 0cm,clip]{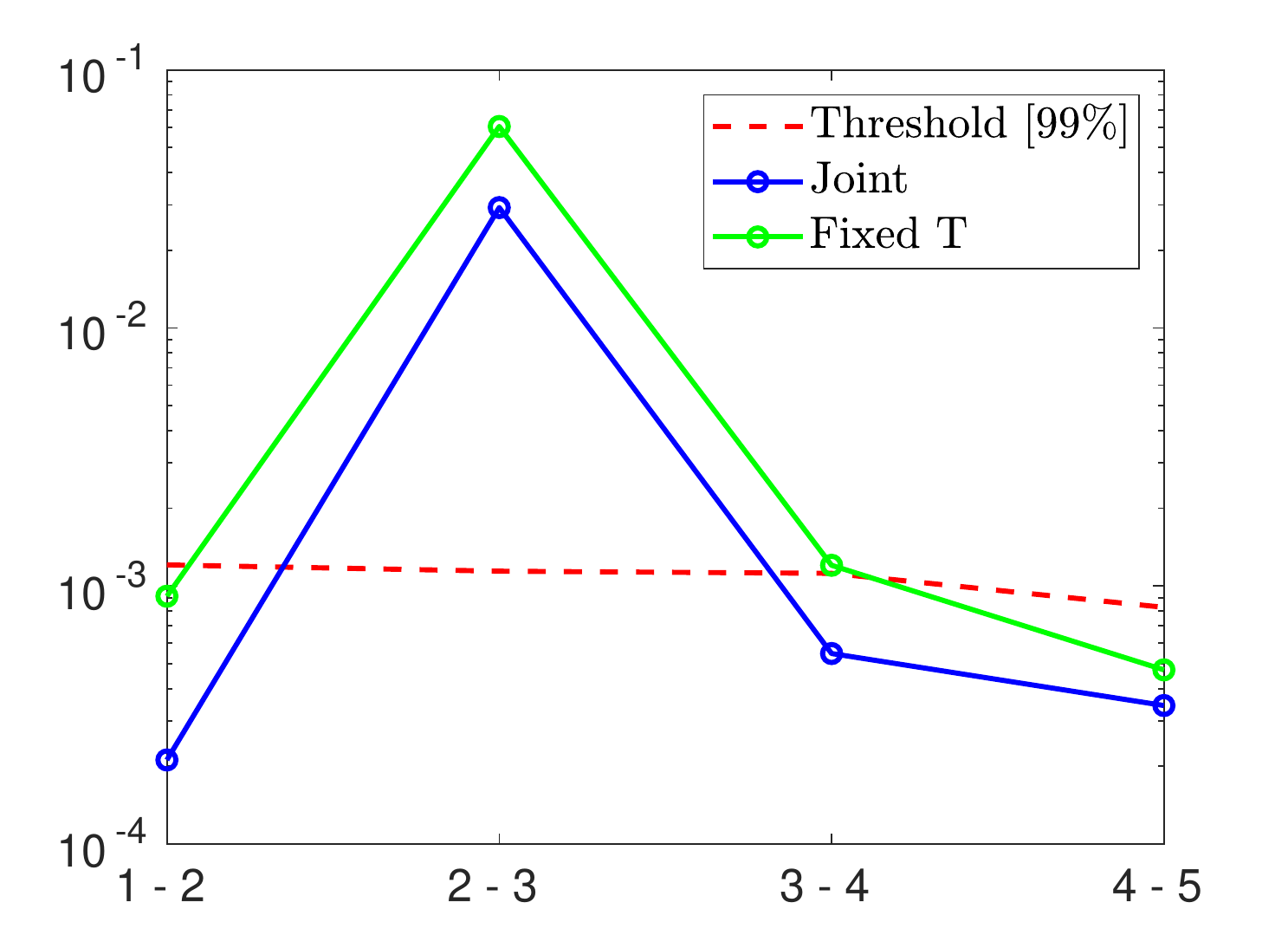}
	\vspace{-3mm}
	\caption{The sum of squared residuals for the merges of different dataset pairs. We see that jointly estimating the transform and 3D points during the merge yields smaller residuals than when estimating the transform before merging as in previous work \cite{flood2019merging}. Change between dataset 2 and 3 is correctly detected.}
	\label{fig_r2d2_change}
\end{figure}

In this experiment we made five separate data collections of a bookshelf. Between collection 2 and 3 we moved an R2D2 figure a few centimeters, see Figure~\ref{fig_r2d2}.
The individual maps were then merged pairwise in sequence --- i.e. 1-2, 2-3, 3-4 and 4-5 --- and compared with our previous work \cite{flood2019merging}
where the transforms between maps were computed prior to the merging. As we see in Figure~\ref{fig_r2d2_change} the residuals are smaller when
jointly estimating merge and transform.
The squared residuals $\tilde{a}$ are then compared with the 99:th percentile of the $\Gamma$ distribution from Section~\ref{sec_hyp_testing}.
We see that change between collection 2 and 3 is correctly detected for both versions, while the previous work with fixed transform is giving a false positive between dataset 3 and 4. Even if the differences are small, this experiment shows that our proposed method performs better than the previous one, despite the problem being harder.

\subsubsection{Experiment in an Office Environment}

\begin{figure}
	\centering
	\begin{tabular}{cc}
		\includegraphics[angle=-15,origin=c,width=4cm,trim = 3cm 4.5cm 3cm 2cm, clip]{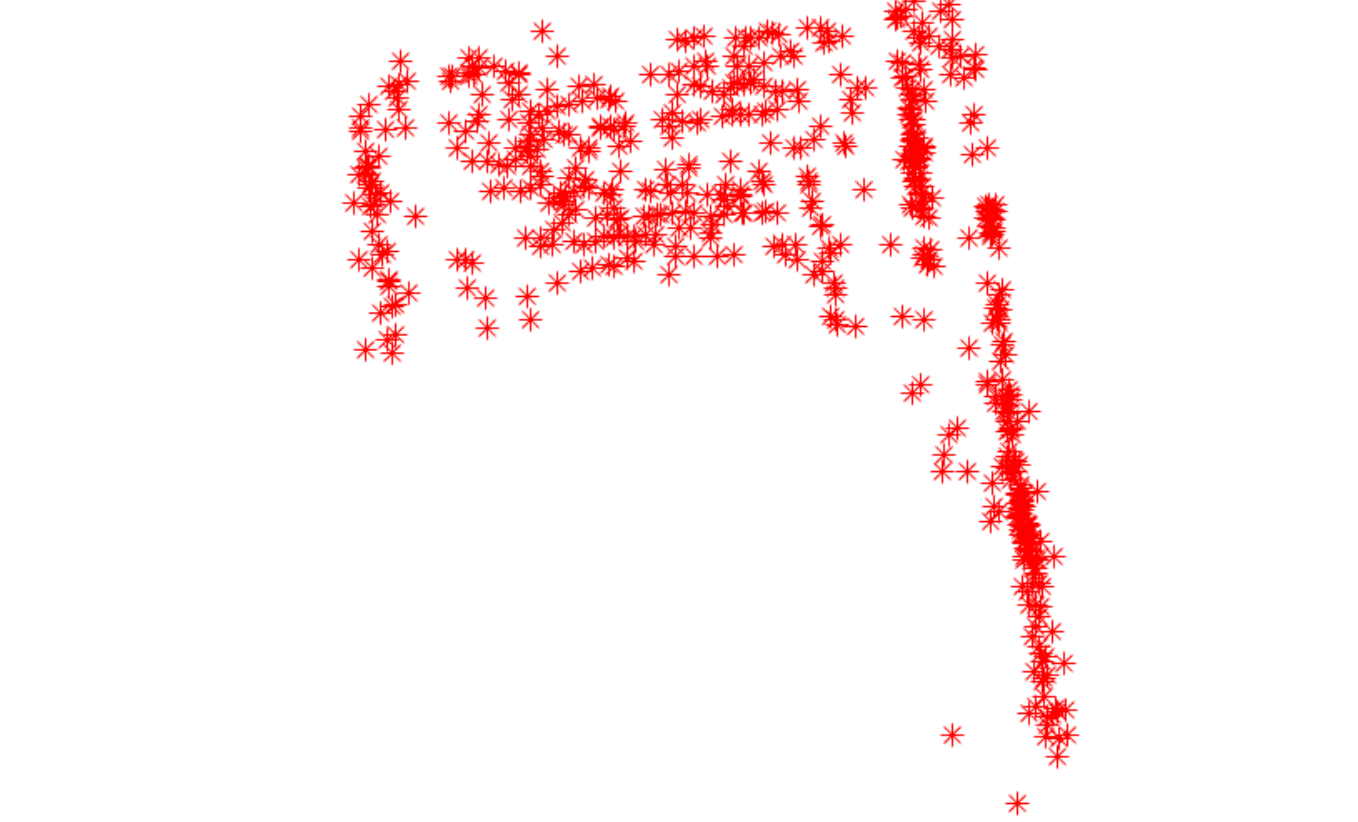} &
		\includegraphics[angle=25,origin=c,width=4cm,trim = 2cm 3.5cm 2cm 2cm, clip]{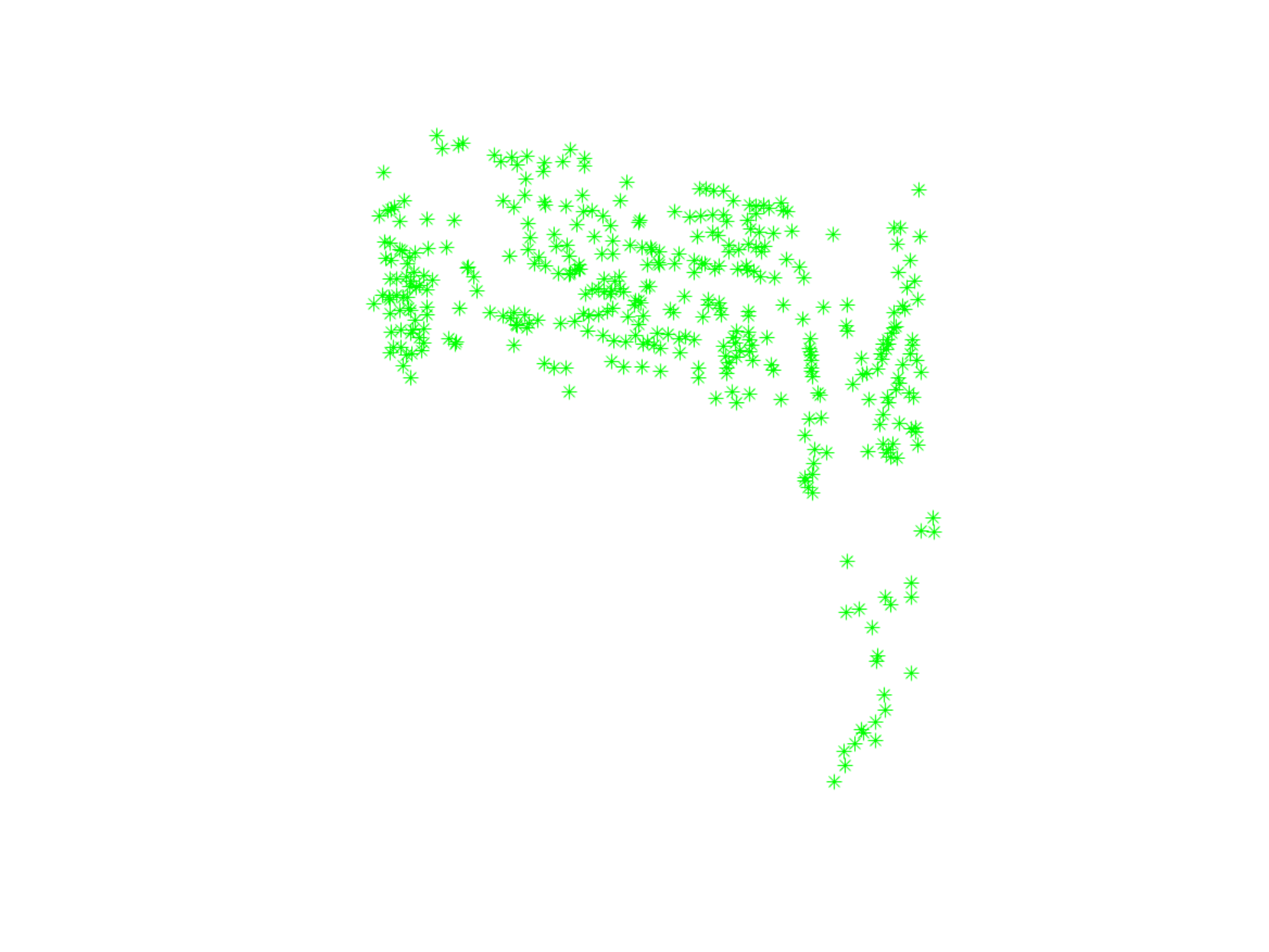} \\
		\includegraphics[width=4cm]{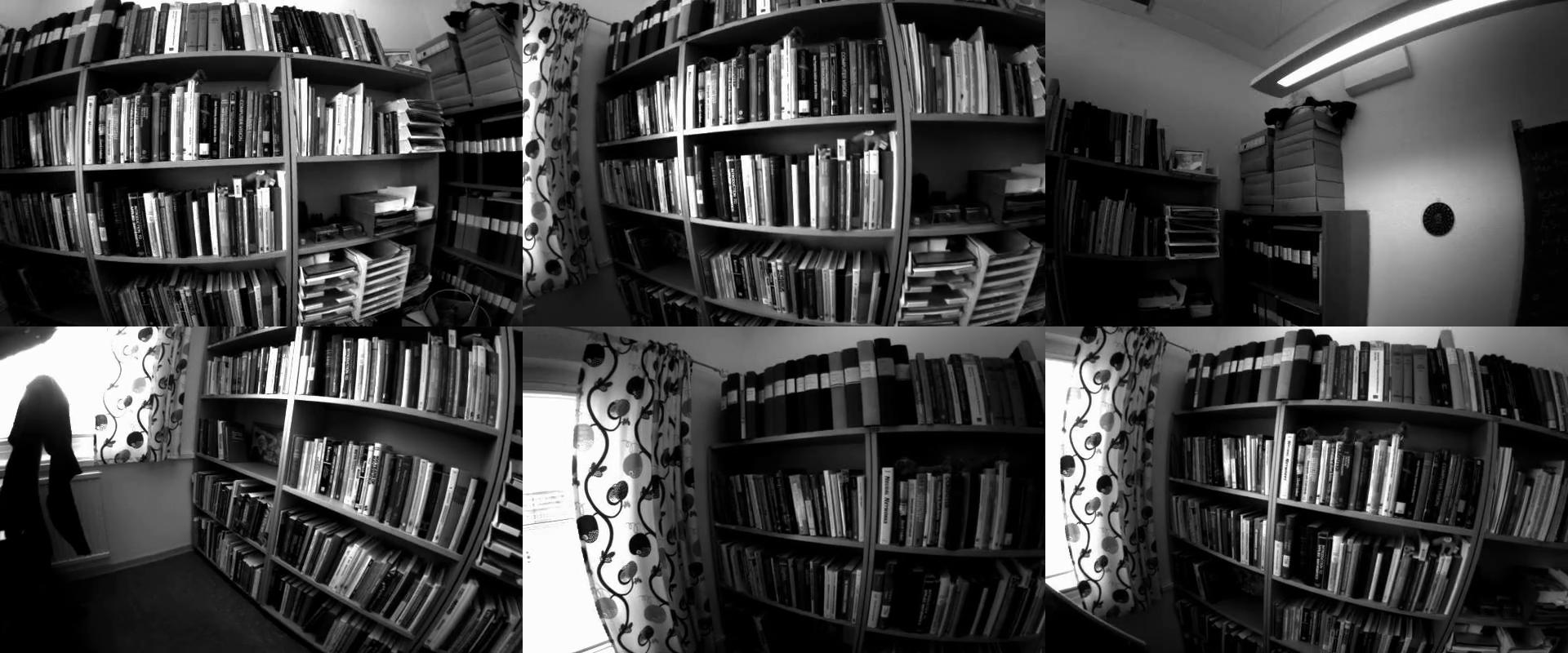} &
		\includegraphics[width=4cm]{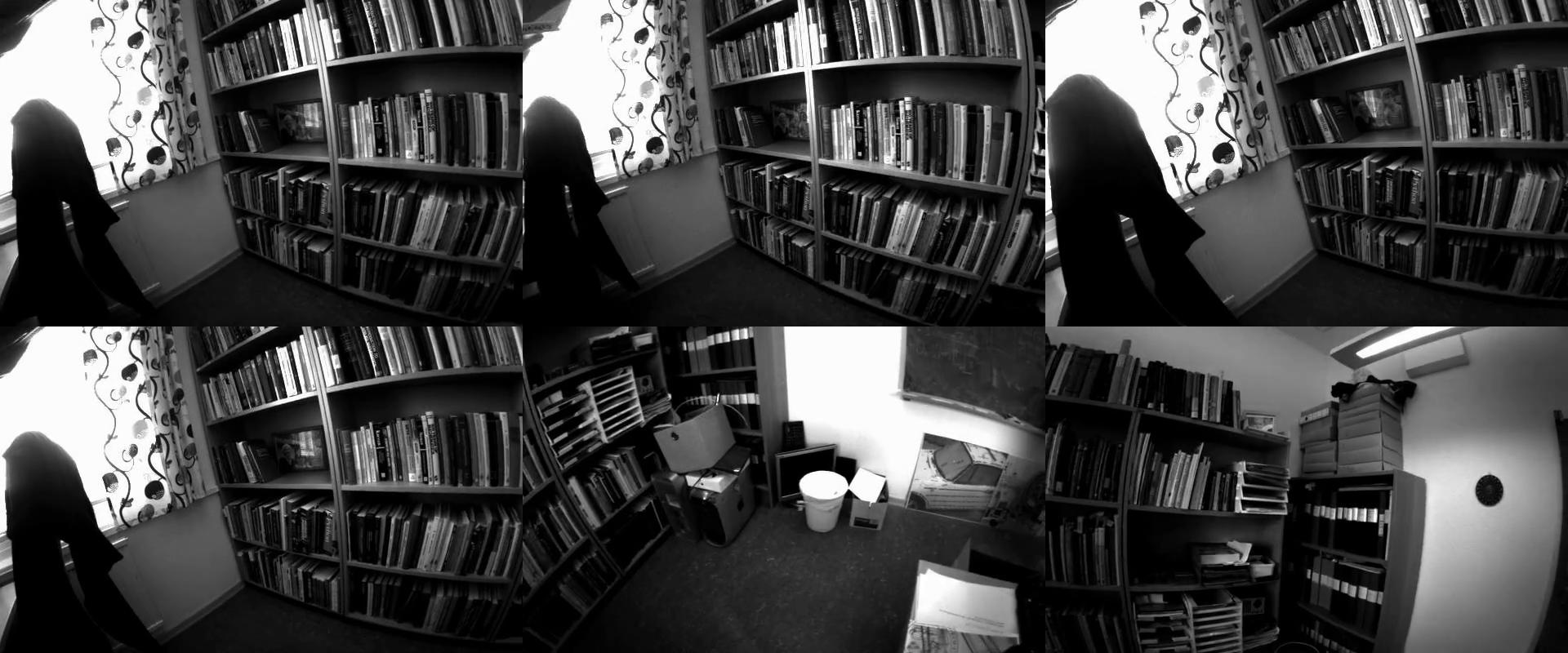} \\
		\includegraphics[width=4cm,trim = 5cm 7.5cm 5cm 1.3cm, clip]{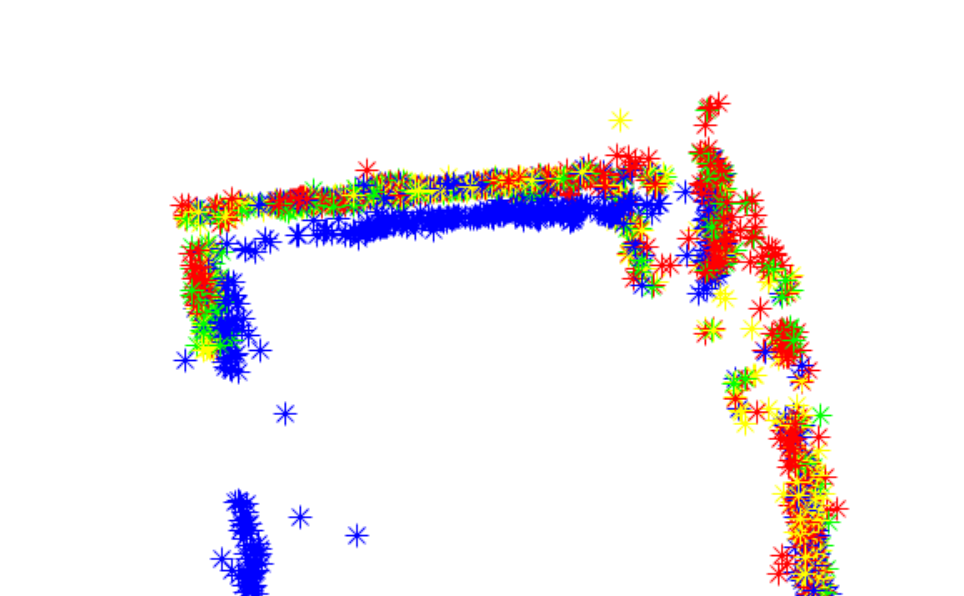} &
		\includegraphics[width=4cm,trim = 5cm 7.5cm 5cm 1cm, clip]{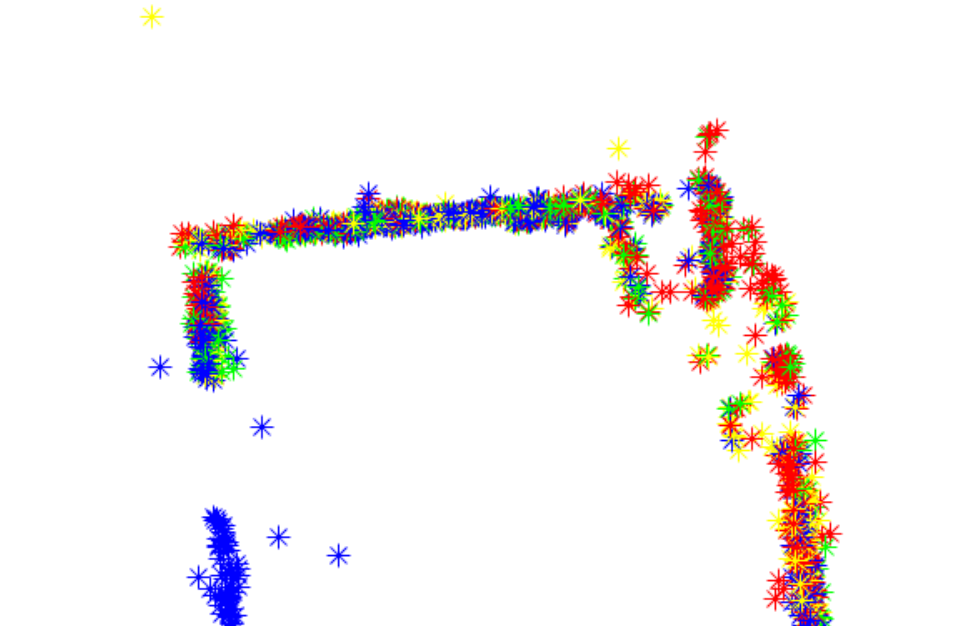} \\
		\end{tabular}
	\caption{The two top rows shows the 3D reconstructions and a few images from two of the four drone recording in the office experiment. The bottom row shows parts of the merged map using Procrustes to the left and our proposed method to the right. Note that the top and left walls are doubled after the Procrustes registration, while our method solves that problem.}
	\label{fig_kallesrum1}
\end{figure}

In the following experiment we made four separate data collections using a drone. Sample images from these datasets, as well as 3D reconstructions, are shown in the two top rows of Figure~\ref{fig_kallesrum1}. Each recording consisted of approximately a minute worth of video footage. The recordings were made with a small drone equipped with a monochrome global
shutter camera (OV9281) with resolution $480 \times 640$ and an
inertial measurement unit (MPU-9250). The 3D reconstructions were generated by a SLAM system built on ORB features \cite{rublee2011orb} and IMU data \cite{forster2015imu}, where the matches are filtered using the technique from \cite{vogiatzis2011video} and the solution is optimised using \cite{kaess2012isam2}. For each of the reconstructions, the object points were saved along with extracted feature locations and descriptors. The feature locations were undistorted prior to saving, to remove fish-eye effects.

The statistics for the four experiments are shown in Table~\ref{table_kallesrum}. The saved descriptors were used to generate hypothesis matches between the different reconstructions. These tentative matches were then tested in a hypothesis and testing framework using the hypothesis test proposed in Section~\ref{sec_hyp_testing} of this paper. This process produced 24 points that were matched across the four experiments.

In the bottom row of Figure~\ref{fig_kallesrum1} we show parts of the merged map after Procrustes to the left and merging using our method to the right. Notice that the top and left wall in the upper left corner had double representations after Procrustes. After merging the two copies of the walls they are positioned on top of each other. After the merge with the proposed method it was possible to identify an additional 346 points that could be merged. 

To validate the performance we selected a few points in one of the maps 
and calculated a number of interpoint distances before and after merging. We also measured these distances in reality with a measuring tape. The results are presented in Table~\ref{table_interpoint_distances}. The results show that our method reduces the error in all the measured distances.


\begin{table}
\caption{Four datasets were collected by drone recordings. The number of 3D points and the size of the Jacobians for each dataset are shown. The proposed method makes it possible to compress the data to a $72 \times 72$ matrix for each dataset. \label{table:kallesrum}}
\label{table_kallesrum}
	\centering
	\begin{tabular}{lllll}
	Bundle	&\# points & Size of full& Size of  compressed\\
	session	&	&  Jacobian &   Jacobian \\
	\hline
1& 999 & $18\, 918 \times 3\, 621$ & $72 \times 72$ \\
2 & 603 & $11\, 972 \times 2\, 151$ & $72 \times 72$ \\
3 & 549 & $11\, 114 \times 1\, 989$ & $72 \times 72$ \\
4 & 386 & $7\, 596 \times 1\, 452$ & $72 \times 72$ \\
merge & 2465 & $49\, 600 \times 8\, 997$ & $ 288 \times 100 $\\
	\end{tabular}
\end{table}

\begin{table}
\caption{Interpoint distances between a few selected points in the office experiment before and after merging using Procrustes registration followed by averaging and our proposed method. The column to the right shows the ground truth distances. \label{table_interpoint_distances}}
%
%
\begin{tabular}{llllll}
Pt 1 & Pt 2 & Dist (mm) & Dist (mm) & Dist (mm) & Dist (mm)\\
ind & ind & one map & merge Pro. & merge our & gt \\
\hline
52 & 766 & 365 & 365 & 220 & 213 \\
52 & 839 & 589 & 589 & 512 & 516 \\
52 & 840 & 1358 & 1296 & 1264 & 1260 \\
60 & 839 & 825 & 825 & 834 & 840 \\
60 & 840 & 879 & 1023 & 860 & 857 \\
\end{tabular}

\end{table}

\section{Conclusion}
In this paper we have presented a new method for merging of 3D maps. The method relies on a low memory footprint representation of the individual residuals that makes it efficient even for a large amount of image data. By bundling over an approximate error, the size of the Jacobian is reduced with several orders of magnitude compared to doing bundle adjustment over all data at once. Furthermore, the method is robust and flexible in the sense that the individual sub-maps do not have to be in the same coordinate system. Our merging method can be used to add two or several maps at once and also for updating a global map using local map estimates. This can furthermore be used to perform loop closing, which is verified using both simulated and real data. Using a hypothesis test based on a statistical analysis of the error we can analyse whether the merge was successful and discover if changes has occurred in the scene between the mappings. In the future we would like to use this to develop a system that can divide a large map into several sub-maps in order to only add the parts of the map that preserves robustness. Another interesting extension would be to generalise the method to rotation averaging.






\bibliographystyle{IEEEtran}
\bibliography{mybib}
%
%
%

\end{document}